\documentclass[runningheads]{llncs}
\usepackage{graphicx}
\usepackage{subfigure}
\usepackage{comment}
\usepackage{array}
\usepackage{multirow}
\usepackage{amsmath,amssymb} 
\usepackage{color} 
\usepackage{verbatimbox}
\usepackage{diagbox}
\usepackage{lipsum}
 
\newcommand\blfootnote[1]{
  \begingroup
  \renewcommand\thefootnote{}\footnote{#1}
  \addtocounter{footnote}{-1}
  \endgroup
}

\begin{document}

\pagestyle{headings}
\mainmatter
\def\ECCVSubNumber{3315}  

\title{Face Anti-Spoofing via Disentangled Representation Learning} 

\titlerunning{Face Anti-Spoofing via Disentangled Representation Learning}

\author{
Ke-Yue Zhang\inst{1,2}$^*$ \and
Taiping Yao\inst{2}$^*$ \and
Jian Zhang\inst{2} \and
Ying Tai\inst{2}$^\dagger$ \and
Shouhong Ding\inst{2} \and
Jilin Li\inst{2} \and
Feiyue Huang\inst{2} \and
Haichuan Song\inst{1} \and
Lizhuang Ma\inst{1} }

\authorrunning{Ke-Yue Zhang, Taiping Yao, et al.}

\institute{ 
East China Normal University, Shanghai, China \and
Youtu Lab, Tencent, Shanghai, China \\
\email{51184501178@stu.ecnu.edu.cn;}
\email{hcsong@cs.ecnu.edu.cn;}
\email{lzma@sei.ecnu.edu.cn;} \\
\email{\{taipingyao,timmmyzhang,yingtai,ericshding,jerolinli,garyhuang\}@tencent.com}}

\maketitle

\begin{abstract}
Face anti-spoofing is crucial to security of face recognition systems. 
Previous approaches focus on developing discriminative models based on the features extracted from images, which may be still entangled between spoof patterns and real persons.
In this paper, motivated by the disentangled representation learning, we propose a novel perspective of face anti-spoofing that disentangles the liveness features and content features from images, and the liveness features is further used for classification. 
We also put forward a Convolutional Neural Network (CNN) architecture with the process of disentanglement and combination of low-level and high-level supervision to improve the generalization capabilities. 
We evaluate our method on public benchmark datasets and extensive experimental results demonstrate the effectiveness of our method against the state-of-the-art competitors. 
Finally, we further visualize some results to help understand the effect and advantage of disentanglement. 

\end{abstract}

$\blfootnote{* equal contribution.}$
$\blfootnote{$^\dagger$ corresponding author.}$

\section{Introduction}
With superior performance than human, face recognition techniques are widely used in smart devices, access control and security scenarios. 
However, the associated safety issues raise concern of public since the accessment of human face is low-cost and a well-designed makeup can easily fool this biometric mechanism. 
These face spoofs, also called Presentation Attacks (PA), vary from simpler printed facial images, video replays to more complicated $3$D mask and facial cosmetic makeup. 
Theoretically, face recognition systems are vulnerable to all spoofs without specific defense, which incurs malicious attacks of hackers, but also encourages the boosting of robust face anti-spoofing algorithm.

Since the primary facial spoof images or videos contain artifacts, researchers put forward several methods based on texture analysis. 
Some handcrafted features are combined with anti-spoofing algorithms, such as Local Binary Pattern(LBP)~\cite{boulkenafet2015face,de2012lbp,de2013can,LBP04}, Histogram of Oriented Gridients(HOG)~\cite{HoG01,HoG02}, Scale Invariant Feature Transform(SIFT)~\cite{SIFT01}, \textit{etc}. 
These cue-based methods use handcrafted features to detect the motion cues such as lip movement or eye blinking for authentication. 
However, these methods couldn't deal with the replay attacks with high-fidelity. 
Recently, Convolutional Neural Network(CNN)-based methods have achieved great progress in face anti-spoofing~\cite{FirstCNN,VGG02,VGG01}. Basically, these methods treat the security issue as a binary classification problem with softmax loss. 
However, they are lack of generalization capability for overfitting on the training dataset. 
Despite many methods use auxiliary information (\textit{i.e.}, facial depth map, rppg signals, \textit{etc}.) to further guide the network in telling the difference between real and spoof~\cite{DeNoise,BASN,RPPG}, these pre-defined characteristics are still insufficient for depicting the authentic abstract spoof patterns since exhausting all possible constraints is impossible.

\begin{figure}[t!]
    \centering
    \includegraphics[width=120mm]{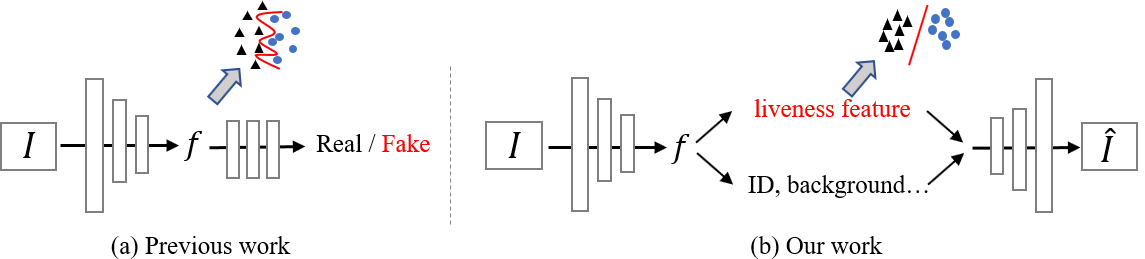}
    \setlength{\belowcaptionskip}{-12pt}
    \caption{\small \textbf{Comparison between previous entangled framework and our disentangled framework.} Previous works learn entangled features which are easily overfitting to the training dataset. In contrast, our disentangled framework distills the liveness features with proper constraints and supervision.}
    \label{illustration}
\end{figure}

Thus the crucial step of face anti-spoofing does not lie in how to precisely pre-define the spoof patterns, but how to achieve the spoof patterns from high-dimensional extracted representations. 
One possible solution is disentangling representations into separate parts.
In disentangle learning~\cite{elegant,MUNIT}, it's a consensus that high-dimensional data can be explained by substantially lower dimensional and semantically meaningful latent representation variables. 
While in face anti-spoofing, the spoof patterns can be viewed as one kind of attributes of face, not just a certain irrelevant noise type or the combination. 
Hence, the problem is transformed into how we can directly target to the liveness information from all the variations of facial images.

As shown in Fig.~\ref{illustration}, we propose a novel disentangled face anti-spoofing technique via separating the latent representation. 
Motivated by~\cite{MUNIT}, we assume the latent space of facial images can be decomposed into two sub-spaces: liveness space and content space. 
Liveness features corresponds to the liveness-related information, while content features integrate remaining liveness-irrelated information in the input images, such as ID and lighting. 
However, in disentangled learning procedure, there exist two challenges on missing $1$) corresponding genuine images for spoof images in translation process and vice versa, $2$) clear research about properties of liveness features in face anti-spoofing literature. 

To tackle above challenges, we introduce low-level texture and high-level depth characteristics to further facilitate disentanglement. 
For the first challenge, we adopt a Generative Adversarial Network(GAN)-like discriminator to guarantee the plausibility of the translated images. 
An auxiliary depth estimator is then introduced to ensure that the liveness information has also been exchanged between genuine and spoof images. 
For the second challenge, checking the properties of liveness features is equivalent to making liveness and content features independent in disentangled framework. 
In order to spilt liveness and content space, we encode the translated images to get reconstructed liveness features again. 
With bidirectional reconstruction loss on images and latent codes, liveness features of diverse spoof patterns are thoroughly extracted in a self-supervised way. 
To further regularize liveness space, we introduce a novel LBP map supervision.
Finally, the spoof classification could be solved in a smaller and more discriminative liveness feature space. 
Hence, our architecture is more likely to achieve good generalization capability.

To sum up, the contributions of this work are three-fold:

$\bullet$ We address face anti-spoofing via disentangled representation learning, which separates latent representation into liveness features and content features.

$\bullet$ We combine low-level texture and high-level depth characteristics to regularize liveness space, which facilitates disentangled representation learning.

$\bullet$ Abundant experiments and visualizations are presented to reveal the properties of liveness features, which demonstrates the effectiveness of our method against the state-of-the-art competitors.

\section{Related Work}
Our method introduces disentangled representation learning to solve face anti-spoofing. Previous related work lies in two perspectives: face anti-spoofing and attributes disentanglement.

\subsubsection{Face Anti-spoofing.}
Early researches focused on hand-crafted feature descriptors, such as LBP~\cite{boulkenafet2015face,de2012lbp,de2013can,LBP04}, HOG~\cite{HoG01,HoG02}, SIFT~\cite{SIFT01} and SURF~\cite{SURF01}, to project the faces into a low-dimension feature space, where traditional classifiers such as SVM are utilized for judgement. 
{There are also some methods using information from different domains, such as HSV and YCrCb color space~\cite{boulkenafet2015face,boulkenafet2016face}, temporal domain~\cite{temporal01,temporal02,temporal03,temporal04}, and Fourier spectrum \cite{Fourier01}.} However, these hand-crafted feature-based methods cannot achieve high accuracy due to limited representation capacity. 

With the rise of deep learning, researchers attempted to tackle the face anti-spoofing with CNN-based features. Initially,~\cite{FirstCNN,VGG02,VGG01} treated the task as a binary classification problem with softmax loss. 
Compared to hand-crafted features, such models gained higher accuracy in intra-testing settings. However, due to the overfitting on training data, their generalization ability are relatively poor. In order to improve the generalization ability, many methods attempted to utilize auxiliary supervision to guide networks. \cite{RPPG} attempted to guide networks with auxiliary supervision of facial depth information and remote-photoplethysmo-graphy (r-ppg) signal. \cite{DeNoise} utilized the spoof images to estimate the spoofing-relevant noise pattern. \cite{Domain} adopted the strategy of domain generalization to achieve improvements in cross-testing. These auxiliary supervision indeed improve generalization. However these methods all handle this problem in the whole feature space, which is disturbed by irrelevant factors.

\subsubsection{Disentangled Representation.}
The key intuition about disentangling is that disentangled representation could factorize the data into distinct informative factors of variations~\cite{challenging}. \cite{VAE03,Infogan} aimed to learn disentangled representations without supervision. \cite{elegant} divided latent features of an facial image into different parts, where each part encodes a single attribute. \cite{MUNIT} assumed that latent space of images can be decomposed into a content space and a style space.

These works inspire us decompose the features of an facial image into content features and liveness features. In face anti-spoofing, content features correspond to the liveness-irrelated information in the images, such as ID, background, Scene lighting, \textit{etc}. On the contrary, liveness features are the key to distinguishing between real persons and attacks. Obviously, we could tackle the face anti-spoofing in the liveness feature space. However, there are many challenges in disentangled learning procedure, such as without the ground truth of the recombined images, diverse styles of spoof, \textit{etc}. In this paper, we combine low-level texture and high-level depth characteristics to facilitate disentangled representation learning.

\section{Disentanglement Framework}
\begin{figure}[t!]
    \centering
    \includegraphics[width=120mm]{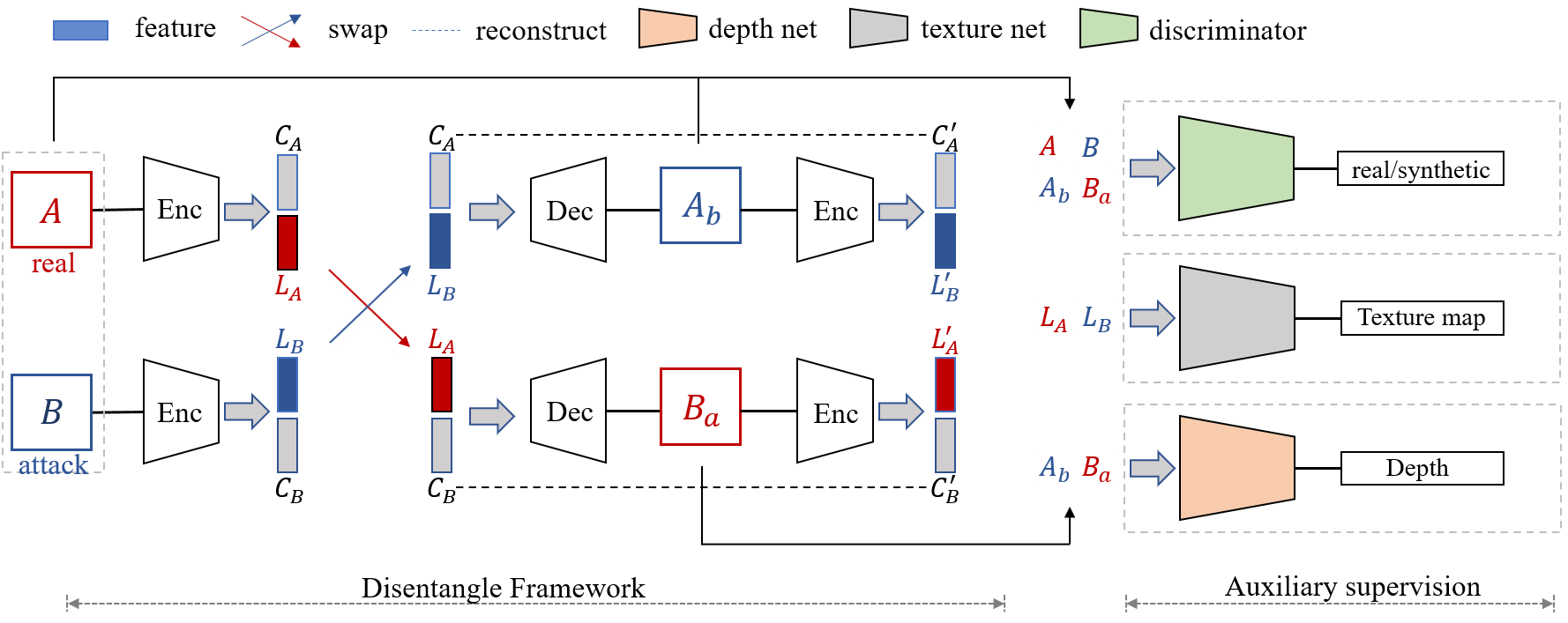}
    \setlength{\belowcaptionskip}{-7pt}
    \caption{\textbf{Overview of our disentanglement framework}. The features of an image are divided into two parts, content features and liveness features. By exchanging the liveness features of the real person and the attack, we can get different reconstructed images with the same content but their liveness attributes are changed. Texture net, depth net and discriminator are proposed to facilitate disentangled representation learning.
}
    \label{structure}
\end{figure}

Our framework mainly consists of two parts: the disentanglement process and the auxiliary supervision. As the core component of our framework explained in Sec.\ref{Disentanglement Process}, the disentanglement process separates the representation into two independent factors, which are liveness features and content features, respectively. As illustrated in Sec.\ref{Auxiliary Supervision}, depth, texture, and discriminative constraints are utilized as auxiliary supervision. By introducing these three auxiliary nets, we consolidate liveness features and further facilitate the disentanglement process. Fig.\ref{structure} illustrates the overview of our method and the entire learning process.

\subsection{Disentanglement Process}
\label{Disentanglement Process}
Disentanglement process is designed to separate liveness features and content features by exchanging and recombining these two features. Inputs of the disentanglement part are two unpaired images $A$ and $B$, where $A$ is randomly chosen from live face images and $B$ is chosen from spoof images. In the encoder part, we first use a convolution block to extract latent code ${Z}$ from inputs. And then two independent convolutional sub-networks encode latent code ${Z}$ into liveness features ${L}$ and content features ${C}$ respectively. This specific structure separates two features from convolving with each other. According to the above process, we can get ${L_A}$, ${C_A}$ and ${L_B}$, ${C_B}$ respectively by encoding images $A$, $B$. Then, we exchange the liveness part ${L_A}$ and ${L_B}$ to obtain images $A_b$ and $B_a$.
\begin{equation}
    A_b = Dec({C_A}, {L_B}), B_a = Dec({C_B}, {L_A}).
\end{equation}
Because the liveness features determine the liveness attributes of the image, we suppose that $A_b$ is a spoof version of image $A$, and $B_a$ is a genuine version of image $B$.
To better decode the latent code back into images, the architecture we used for the decoder is symmetrical with the encoder. Besides, following the U-Net~\cite{UNet} structure, the shortcuts are added from the middle layers in encoder to the corresponding layers in decoder to bring the original information as an auxiliary context for improving visual quality. 
To further guarantee that liveness information and content information can be split completely, we encode images $A_b$, $B_a$ again to get ${C_A'}, {L_B'}$ and ${C_B'}, {L_A'}$, and introduce a bidirectional reconstruction loss~\cite{MUNIT} to encourage reconstruction in two sequential processes (\textit{i.e.}, from images to images and from latent features to latent features).

\noindent\textbf{Image Reconstruction.}
The combination of the encoder and decoder should be capable of reconstructing any image $x_i$ from the datasets:
\begin{equation}
\begin{split}
    {\mathcal{L}_{x_i}^{rec}} 
    = \mathbb{E}_{x_i \sim p(x_i)} \left\| D(E(x_i)) - x_i \right\|_{1} ,
\end{split}
\end{equation}
where $p(x_i)$ is the distribution of original images in the datasets, $E$ is the encoder and $D$ is the decoder.

\noindent\textbf{Latent Reconstruction.}
Given a pair of liveness features and content features at translation time, we should be able to reconstruct it after decoding and encoding.
\begin{equation}
\begin{split}
    {\mathcal{L}_{z_i}^{rec}} 
    = \mathbb{E}_{z_i \sim q(z_i)} \left\| E(D(z_i)) - z_i \right\|_{1}
    \label{latent_reconstruction}
\end{split}
\end{equation}
where $z_i$ is the combination of liveness features $L_i$ and content features $C_i$, and $q(z_i)$ is the distribution of latent code.

\subsection{Auxiliary Supervision}
\label{Auxiliary Supervision}

In this section, we introduce three auxiliary supervision: LBP map, depth map and discrimination supervision, which promote the disentanglement process collaboratively. Discrimination supervision ensures the visual quality of generated image. Depth and LBP supervision are plugged into different parts to guarantee the generated image being in correct category when their liveness features are exchanged. The LBP map and depth map together regularize the liveness feature space, making it the key factor to distinguish between real persons and spoof patterns.  
{The detailed structure of three auxiliary nets are illustrated in Tab.~\ref{detail_net}. Each convolutional layer is followed by a batch normalization layer and a Rectified Linear Unit (ReLU) activation function with $3 \times 3$ kernel size.}

\begin{table}[t!]
  \centering
  \scriptsize
  \caption{\textbf{The details of Auxiliary nets of our method}.}
    \linespread{1}\selectfont
    \begin{tabular}{ccc||ccc||ccc}
    \hline
    \hline
    \multicolumn{3}{c||}{LBP Net} & \multicolumn{3}{c||}{Depth Net} & \multicolumn{3}{c}{Discriminator} \\
    \hline
    Layer & chan./Stri. & Out.Size & Layer & chan./Stri. & Out.Size & Layer & chan./Stri. & Out.Size \\
    \hline
    \multicolumn{3}{c||}{\textbf{Input}:liveness features} & \multicolumn{3}{c||}{\textbf{Input}:image} & \multicolumn{3}{c}{\textbf{Input}:image} \\
    \hline
            &      &     & conv2-0 & 64/1  & 256 &         &      &      \\
    \hline
    conv1-0 & 384/1 & 32 & conv2-1 & 128/1 & 256 & conv3-1 & 64/1 & 256 \\
            &      &     & conv2-2 & 196/1 & 256 & pool3-1 & -/2  & 128 \\
            &      &     & conv2-3 & 128/1 & 256 &         &      &     \\
            &      &     & pool2-1 & -/2   & 128 &         &      &     \\
    \hline
    conv1-1 & 128/1 & 32 & conv2-4 & 128/1 & 128 & conv3-2 & 128/1 & 128 \\
            &      &     & conv2-5 & 196/1 & 128 & pool3-2 & -/2  & 64 \\
            &      &     & conv2-6 & 128/1 & 128 &         &      &     \\
            &      &     & pool2-2 & -/2   & 64  &         &      &     \\
    \hline
    conv1-2 & 64/1 & 32  & conv2-7 & 128/1 & 64 & conv3-3 & 256/1 & 64 \\
            &      &     & conv2-8 & 196/1 & 64 & pool3-3 & -/2  & 32 \\
            &      &     & conv2-9 & 128/1 & 64 &         &      &     \\
            &      &     & pool2-3 & -/2   & 32 &         &      &     \\
    \hline
    \multicolumn{3}{c||}{conv1-2} & \multicolumn{3}{c||}{pool2-1+pool2-2+pool2-3} & \multicolumn{3}{c}{vectorize} \\
    \hline
            &      &     & conv2-10 & 128/1 & 32 &         &       &     \\
            &      &     & conv2-11 & 64/1  & 32 &         &       &     \\
    conv1-3 & 1/1  & 32  & conv2-12 & 1/1   & 32 & fc3-1   & 1/1   & 2  \\
    \hline
    \hline
    \end{tabular}
  \label{detail_net}
\end{table}

\subsubsection{Texture Auxiliary Supervision.}
Liveness features are the essential characteristic of a face image, which determine the liveness categories of the image. Thus when swapping liveness features between a real person and an attack, categories of images and estimated depth maps should be changed simultaneously. 
And the estimated depth map is usually considered to be related to factors such as facial lighting and shadows, which are contained in the texture information of the face. What's more, previous works have proven that texture is an important clue in face anti-spoofing. Therefore, LBP map is adopted to regularize the liveness features in disentanglement framework. 
{Although LBP features contain some additional information, proposed disentanglement framework utilize Latent Reconstruction Loss to constrain liveness features to learn only essential information. }
To make the features distinctive, for the genuine faces, we use the LBP map extracted by the algorithm in~\cite{LBP} as texture supervision. While for the spoof face, a zero map serves as the ground truth.
\begin{equation}
\begin{split}
    {\mathcal{L}_{lbp}} &= 
    \mathbb{E}_{l_i \sim P(l_i), x_i \sim P(x_i)} \left\| LBP(l_i) - lbp_{x_i} \right\|_{1}  \\
     &+ \mathbb{E}_{l_i \sim N(l_i), x_i \sim N(x_i)} \left\| LBP(l_i) - \mathbf{0} \right\|_{1}  
\end{split}
\end{equation}
where $LBP$ is the LBP Estimator Net, $P(x_i)$ is the distribution of live face images in the datasets, $P(l_i)$ is the distribution of liveness space of live face images, $N(x_i)$ is the distribution of spoof images in the datasets, $N(l_i)$ is the distribution of liveness space of spoof images, $lbp_{x_i}$ means the lbp map of live face images $x_i$ and $\mathbf{0}$ means the zero maps for spoof images.

\subsubsection{Depth Supervision.}
Depth map is commonly used as an auxiliary supervision in face anti-spoofing tasks. In our disentanglement framework, we combine LBP map and depth map supervision to regularize the liveness feature space. 
Similarly as LBP branch, we use pseudo-depth as ground truth for live face images and zero map for spoof images. The pseudo-depth is estimated by the 3D face alignment algorithm in~\cite{Depth}. During training stage, depth net only provides the supervision and does not update parameters. Since the reconstructed images $A'$ and generated $B_a$ are live images, and the reconstructed images $B'$ and generated $A_b$ are spoof images, the corresponding depth map of above images should be the depth of the face in images $A$, $B$ and two zero maps. Then the loss of depth is formulated as:  

\begin{equation}
\begin{split}
    {\mathcal{L}_{dep}} &= 
    \mathbb{E}_{x_i \sim N(x_i)} \left\| Dep(x_i) - \mathbf{0} \right\|_{1} + \mathbb{E}_{x_i \sim P(x_i)} \left\| Dep(x_i) - dep_{x_i} \right\|_{1}
\end{split}
\end{equation}
where $Dep$ is the parameters fixed depth net, $P(x_i)$ is the distribution of live face images, $N(x_i)$ is the distribution of spoof images, $dep_{x_i}$ is the depth map of live face images $x_i$ and $\mathbf{0}$ means the zero maps for spoof images correspondingly.

\subsubsection{Discriminative Supervision.}
For ensuring the visual plausibility of generated images, we apply discriminative supervision on the generated images. Discriminative supervision is used for distinguishing between the generated images ($A', B', A_b, B_a$) and the original images ($A, B$). At the same time, disentanglement framework aims to produce plausible images which would be classified as non-synthetic images under discriminative supervision. Nevertheless, the receptive field of a single discriminator is limited for large images. We use multi-scale discriminators~\cite{MultiDis} to address this problem. Specifically, we deploy two
identical discriminators with varied input resolution. 
{The discriminator with a larger input scale is denoted as $D_1$, which guides the disentanglement net to generate finer details. And the other discriminator with a smaller input scale is denoted as $D_2$, which guides the disentanglement net to preserve more global information.} 
In the training process, there are two consecutive steps in each iteration. In the first step, we fix disentanglement net and update the discriminator, 

\begin{equation}
\begin{split}
    {\mathcal{L}_{D}^{Dis}} = &-\mathbb{E}_{I \in R}log(D_1(I)) - \mathbb{E}_{I \in G}log(1-D_1(I))  \\
                          &-\mathbb{E}_{I \in R}log(D_2(I)) - \mathbb{E}_{I \in G}log(1-D_2(I)) 
    \label{dis}
\end{split}
\end{equation}
where $R$ and $G$ are the sets of real and generated images respectively. In the second step, we fix the discriminator and update the disentanglement net,

\begin{equation}    
    {\mathcal{L}_{D}^{Gen}} = - \mathbb{E}_{I \in G}log(D_1(I)) - \mathbb{E}_{I \in G}log(D_2(I))  
\end{equation}

\subsubsection{Loss Function.}
\label{Loss Function}
The final loss function of training process is the weighted summation of the loss functions above,
\begin{equation}    
\begin{split}
    {\mathcal{L}} = 
    {\mathcal{L}_{D}^{Gen}} + \lambda_1{\mathcal{L}_{x_i}^{rec}} + \lambda_2{\mathcal{L}_{z_i}^{rec}} + \lambda_3{\mathcal{L}_{dep}} + \lambda_4{\mathcal{L}_{lbp}}
    \label{gen}
\end{split}
\end{equation}
where $\lambda_1, \lambda_2, \lambda_3, \lambda_4$ are the weights. Following common adversarial training pipeline, we alternately optimize discriminator and disentanglement net. The weights are empirically selected to balance each loss term. 
%We set $\lambda_1$ largest for image reconstruction.

\section{Experimental Results}
\subsection{Experimental Setting}
\textbf{Databases.}
We test our method on four face anti-spoofing databases: Oulu-NPU~\cite{Oulu}, SiW~\cite{RPPG}, CASIA-MFSD~\cite{CASIA} and Replay-Attack~\cite{Replay}. We evaluate our intra-testing performance on Oulu-NPU and SiW datasets, and conduct cross-testing by training on Replay-Attack or CASIA-MFSD and testing on the other.

\noindent\textbf{Metrics.}
To compare with previous works, we report the performance via the following metrics: Attack Presentation Classification Error Rate (\textsl{APCER})~\cite{metrics}, Bona Fide Presentation Classification Error Rate (\textsl{BPCER})~\cite{metrics}, Average Classification Error Rate (\textsl{ACER}) = (\textsl{APCER+BPCER})/2~\cite{metrics} and Half Total Error Rate (\textsl{HTER}) = (False Acceptance Rate + False Rejection Rate)/2~\cite{metrics}.

\begin{table}[t!]
  \centering
  \caption{\textbf{The intra-testing results of four protocols of Oulu-NPU dataset}.}
    \linespread{1}\selectfont
    \begin{tabular}{|c|p{2.25cm}<{\centering}|c|c|c|}
    \hline
    Protocol & Method & APCER(\%) & BPCER(\%) & ACER(\%) \\
    \hline
    \multirow{5}*{1}    & STASN\cite{STASN}      & 1.2   & 2.5            & 1.9 \\ \cline{2-5}
                        & Auxiliary\cite{RPPG}   & 1.6   & 1.6            & 1.6 \\ \cline{2-5}
                        & FaceDe-S\cite{DeNoise} & \textbf{1.2}   & 1.7            & 1.5 \\ \cline{2-5}
                        & FAS-TD\cite{exploiting}& 2.5   & \textbf{0.0}   & \textbf{1.3} \\ \cline{2-5}
                        & Ours                   & 1.7   & 0.8   & \textbf{1.3} \\ \cline{2-5}
    \hline
    \hline
    \multirow{5}*{2}    & Auxiliary\cite{RPPG}    & 2.7   & 2.7   & 2.7 \\ \cline{2-5}
                        & GRADIANT\cite{fusion03} & 3.1   & 1.9   & 2.5 \\ \cline{2-5}
                        & STASN\cite{STASN}       & 4.2   & \textbf{0.3}   & 2.2 \\ \cline{2-5}
                        & FAS-TD\cite{exploiting} & 1.7   & 2.0    & \textbf{1.9} \\ \cline{2-5}
                        & Ours                    & \textbf{1.1}   & 3.6   & 2.4 \\ \cline{2-5}
    \hline
    \hline
    \multirow{5}*{3}    & FaceDe-S\cite{DeNoise} & 4.0$\pm$1.8 & 3.8$\pm$1.2 & 3.6$\pm$1.6 \\ \cline{2-5}
                        & Auxiliary\cite{RPPG}   & 2.7$\pm$1.3 & 3.1$\pm$1.7 & 2.9$\pm$1.5 \\ \cline{2-5}
                        & STASN\cite{STASN}      & 4.7$\pm$3.9 & \textbf{0.9$\pm$1.2} & 2.8$\pm$1.6 \\ \cline{2-5}
                        & BASN\cite{BASN}        & \textbf{1.8$\pm$1.1} & 3.6$\pm$3.5 & 2.7$\pm$1.6 \\ \cline{2-5}
                        & Ours                   & 2.8$\pm$2.2 & 1.7$\pm$2.6 & \textbf{2.2$\pm$2.2} \\ \cline{2-5}
    \hline
    \hline
    \multirow{5}*{4}    & FAS-TD\cite{exploiting}& 14.2$\pm$8.7   & 4.2$\pm$3.8    & 9.2$\pm$6.0 \\ \cline{2-5} 
                        & STASN\cite{STASN}      & 6.7$\pm$10.6 & 8.3$\pm$8.4  & 7.5$\pm$4.7 \\ \cline{2-5}
                        & FaceDe-S\cite{DeNoise} & \textbf{5.1$\pm$6.3}  & 6.1$\pm$5.1  & 5.6$\pm$5.7 \\  \cline{2-5}
                        & BASN\cite{BASN}        & 6.4$\pm$8.6  & \textbf{3.2$\pm$5.3}  & 4.8$\pm$6.4 \\  \cline{2-5}
                        
                        & Ours                   & 5.4$\pm$2.9  & 3.3$\pm$6.0  & \textbf{4.4$\pm$3.0} \\  \cline{2-5}
    \hline
    \end{tabular}
  \label{Oulu-NPU}
\end{table}

\noindent\textbf{Implementation Details.}
All datasets above are stored in video format. We use a face detector or face location files in datasets to crop the face and resize it to 256 $\times$ 256. 
{For each frame, we combine scores of estimated LBP map and Depth map to detect attack for fully utilizing the low-level texture information and high-level global information, as the methods in~\cite{DeNoise}, \textit{i.e.}, $score = (\left\| map_{lbp} \right\| + \left\| map_{depth} \right\|)/2$.}
We implement method in Pytorch~\cite{pytorch}. Models are trained with batch size of 4. In each epoch, we select negative images and positive images with the ratio 1 : 1. To train network, we use learning rate of 1e-5 with Adam optimizer~\cite{adam} and set $\lambda_1$ to $\lambda_4$ in Eqn.~\ref{gen} as 10, 1, 1 and 2. Depth net is pre-trained and remains fixed during the training of other three nets, and all networks are trained with the same data in each protocol.
{In inference stage, reconstruction and translation procedure are both detached, thus the speed of our method is acceptable, which achieves 77.97$\pm$0.18 FPS on GeForce GTX 1080.}

\subsection{Experimental Comparison}
In this section, we show the superiority of disentanglement and further illustrate translation results. To verify the performance of our method, we conduct experiments on Oulu-NPU and SiW for intra-testing results, CASIA and Replay-Attack for cross-testing results. Then we demonstrate some examples to show details of translation, which verifies the validity of the liveness features. 

\begin{table}[t!]
  \centering
  \caption{\textbf{The intra-testing results of three protocols of SiW dataset}.}
    \linespread{1.1}\selectfont
    \begin{tabular}{|c|p{2.25cm}<{\centering}|c|c|c|}
        \hline
        Protocol & Method & APCER(\%) & BPCER(\%) & ACER(\%) \\
        \hline
        \multirow{5}*{1}    & Auxiliary\cite{RPPG}  & 3.58 & 3.58 & 3.58 \\ \cline{2-5}
                            & STASN\cite{STASN}     & - & - & 1.00 \\ \cline{2-5}
                            & FAS-TD\cite{exploiting}  & 0.96 & 0.50 & 0.73 \\ \cline{2-5}
                            & BASN\cite{BASN}       & - & - & 0.37 \\ \cline{2-5}
                            & Ours                  & 0.07 & 0.50 & \textbf{0.28} \\ \cline{2-5}
        \hline
        \hline
        \multirow{5}*{2}    & Auxiliary\cite{RPPG}  & 0.57$\pm$0.69 & 0.57$\pm$0.69 & 0.57$\pm$0.69 \\ \cline{2-5}
                            & STASN\cite{STASN}     & - & - & 0.28$\pm$0.05 \\ \cline{2-5}
                            & FAS-TD\cite{exploiting}  & 0.08$\pm$0.17 & 0.21$\pm$0.16 & 0.15$\pm$0.14 \\ \cline{2-5}
                            & BASN\cite{BASN}       & - & - & 0.12$\pm$0.03 \\ \cline{2-5}
                            & Ours                  & 0.08$\pm$0.17 & 0.13$\pm$0.09 & \textbf{0.10$\pm$0.04} \\ \cline{2-5}
        \hline
        \hline
        \multirow{5}*{3}    & STASN\cite{STASN}  & - & - & 12.10$\pm$1.50 \\ \cline{2-5}
                            & Auxiliary\cite{RPPG}  & 8.31$\pm$3.81 & 8.31$\pm$3.80 & 8.31$\pm$3.81 \\ \cline{2-5}
                            & BASN\cite{BASN}       & - & - & 6.45$\pm$1.80 \\ \cline{2-5}
                            & FAS-TD\cite{exploiting}  & 3.10$\pm$0.79 & 3.09$\pm$0.83 & \textbf{3.10$\pm$0.81} \\ \cline{2-5}
                            
                            & Ours                  & 9.35$\pm$6.14 & 1.84$\pm$2.60 & 5.59$\pm$4.37 \\ \cline{2-5}
        \hline
      \end{tabular}
      \label{SiW}
\end{table}

\subsubsection{Intra-Testing.}
Intra-testing is evaluated on Oulu-NPU and SiW datasets. We utilize the protocols defined in each dataset. Tab.~\ref{Oulu-NPU} shows the comparison of our method with the best four methods on Oulu dataset. Our method achieves better results in protocols 1, 3 and 4, while gets slightly worse ACER in protocol 2. For protocol 4 evaluating all variations in Oulu, our method gets the best results, which verifies that our method has better generalization performance. 
Following~\cite{BASN}, we report the ACER on three protocols of SiW. Tab.~\ref{SiW} shows that our method achieves better results among the frame based methods.

\begin{table}[t!]
  \centering
  \caption{\textbf{The cross-testing results on CASIA-MFSD and Replay-Attack}.}
    \linespread{1.1}\selectfont
    \begin{tabular}{|p{3.00cm}<{\centering}|c|c|c|c|}
        \hline
        \multirow{3}*{Method}  & Train & Test   & Train  & Test   \\ \cline{2-5}
                               & CASIA & Replay & Replay & CASIA  \\
                               & MFSD  & Attack & Attack & MFSD   \\
        \hline
        Motion-Mag\cite{Motion-Mag}             & \multicolumn{2}{c|}{50.1\%} & \multicolumn{2}{c|}{47.0\%} \\ \hline
        Spectral cubes\cite{Spectral-cubes}     & \multicolumn{2}{c|}{34.4\%} & \multicolumn{2}{c|}{50.0\%} \\ \hline
        LowPower\cite{lowpower}                 & \multicolumn{2}{c|}{30.1\%} & \multicolumn{2}{c|}{35.6\%} \\ \hline
        CNN\cite{FirstCNN}                      & \multicolumn{2}{c|}{48.5\%} & \multicolumn{2}{c|}{45.5\%} \\ \hline
        STASN\cite{STASN}                       & \multicolumn{2}{c|}{31.5\%} & \multicolumn{2}{c|}{30.9\%} \\ \hline
        FaceDe-S\cite{DeNoise}                  & \multicolumn{2}{c|}{28.5\%} & \multicolumn{2}{c|}{41.1\%} \\ \hline
        Auxiliary\cite{RPPG}                    & \multicolumn{2}{c|}{27.6\%} & \multicolumn{2}{c|}{\textbf{28.4\%}} \\ \hline
        BASN\cite{BASN}                         & \multicolumn{2}{c|}{23.6\%} & \multicolumn{2}{c|}{29.9\%} \\ \hline
        Ours                                    & \multicolumn{2}{c|}{\textbf{22.4\%}} & \multicolumn{2}{c|}{30.3\%} \\ \hline
        \end{tabular}
        \label{CASIA-Replay}
\end{table}
\subsubsection{Cross-Testing.}
We evaluate the generalization capability by conducting cross-dataset evaluations. Following the related work, CASIA-MFSD and Replay-Attack are used for the experiments and the results are measured in HTER. The results are shown in Tab.~\ref{CASIA-Replay}. For fair comparison, we compare with methods using only single frame information. Our method achieves 1.2 pp lower HTER than the state-of-the-art from CASIA-MFSD to Replay-Attack and gets comparable HTER from Replay-Attack to CASIA-MFSD. This results also prove that our method with disentanglement has better generalization capability.

\subsubsection{Translation Result.}
\label{Translation Result}
\begin{figure}[t!]
    \centering
    \includegraphics[width=120mm]{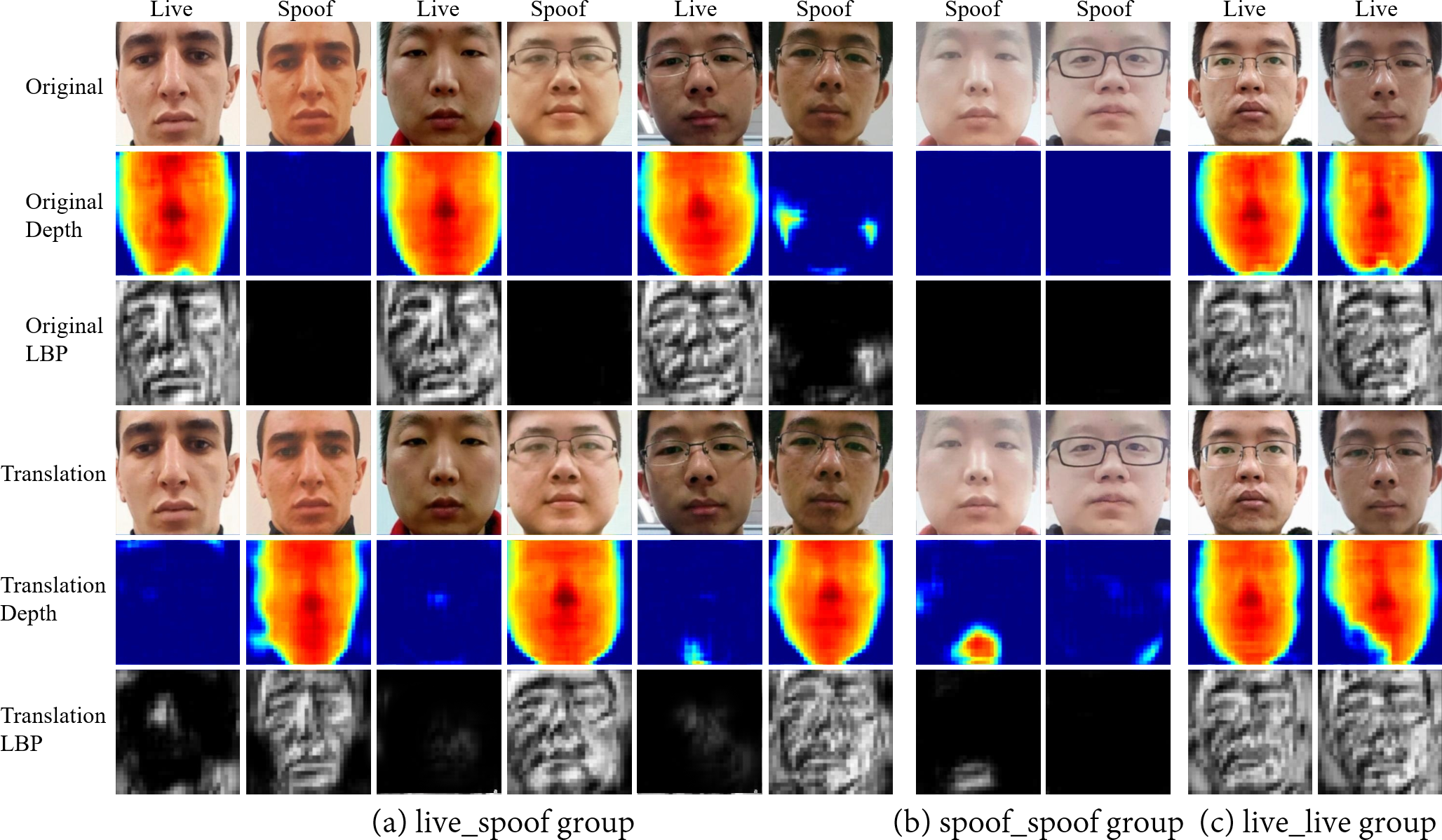}
    \caption{\textbf{Illustrations of translation results with corresponding depth map and LBP map.} We swap liveness features between every two columns. The exchanging of depth and LBP map verifies that liveness features are the key part of live face images.}
    \label{demonstrate}
\end{figure}

\begin{figure}[t!]
    \centering
    \includegraphics[width=120mm]{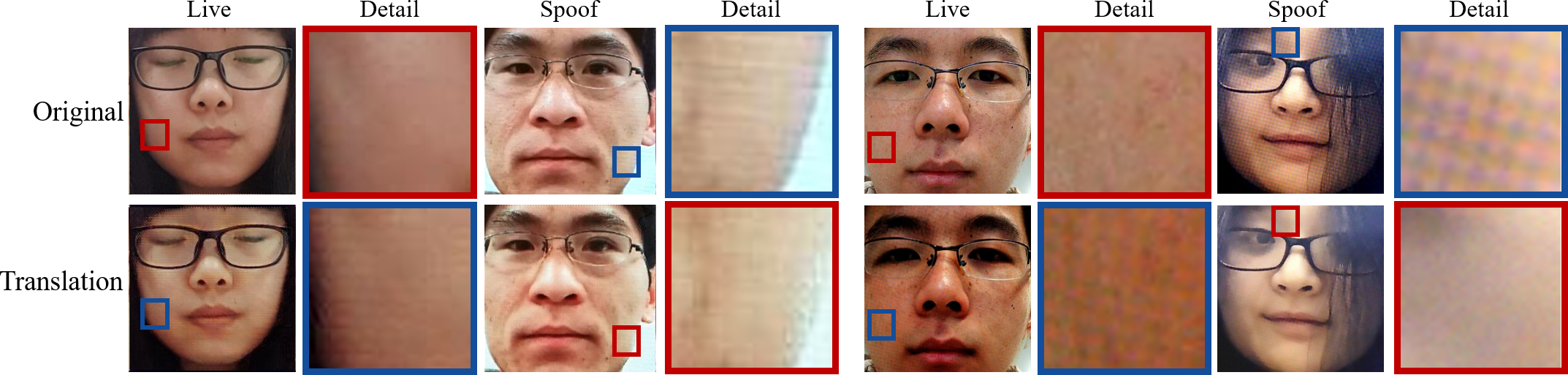}
    \setlength{\belowcaptionskip}{-12pt}
    \caption{\textbf{Illustrations of exchanging live and spoof details}. The first row is the original image, and the second row is the translation results.
    Red rectangular is referred as the details of live images while blue refer to the details of the spoof images. 
    }
    \label{detail}
\end{figure}

We demonstrate some examples of translation from Oulu protocol 1 in three groups: live-spoof, live-live, spoof-spoof, as shown in Fig.~\ref{demonstrate}. In the live-spoof group, depth map changes with the exchange of the liveness features. While in live-live group and spoof-spoof group, the liveness features changing doesn't result in the change of depth map, which implies that liveness features indeed determine whether the image is live. The difference between each two columns of live face and spoof images is \textbf{light, ID, background} respectively. As the translation shows, there are no changes about these factors with the category changing, which means that liveness features do not contain these factors. 
Fig.~\ref{detail} shows two sets of live and attack images and their local area details. As shown in the figure, there is a big difference between the local details of the real person and the attack, and the attack images often have some repetitive streaks. And after combining the liveness features from the attack images, the local details of the translation results are similar to the corresponding attacks, which shows that the liveness features have not only learned the difference between real people and attacks, but also learned different attack details.

\subsection{Ablation Study}
To study the effect of disentanglement, different supervision and score fusion methods, we conduct ablation experiments on Oulu-NPU protocol 1 respectively.

\subsubsection{Liveness Feature Distribution.}
\begin{figure}[t!]
    \centering
    \includegraphics[width=120mm]{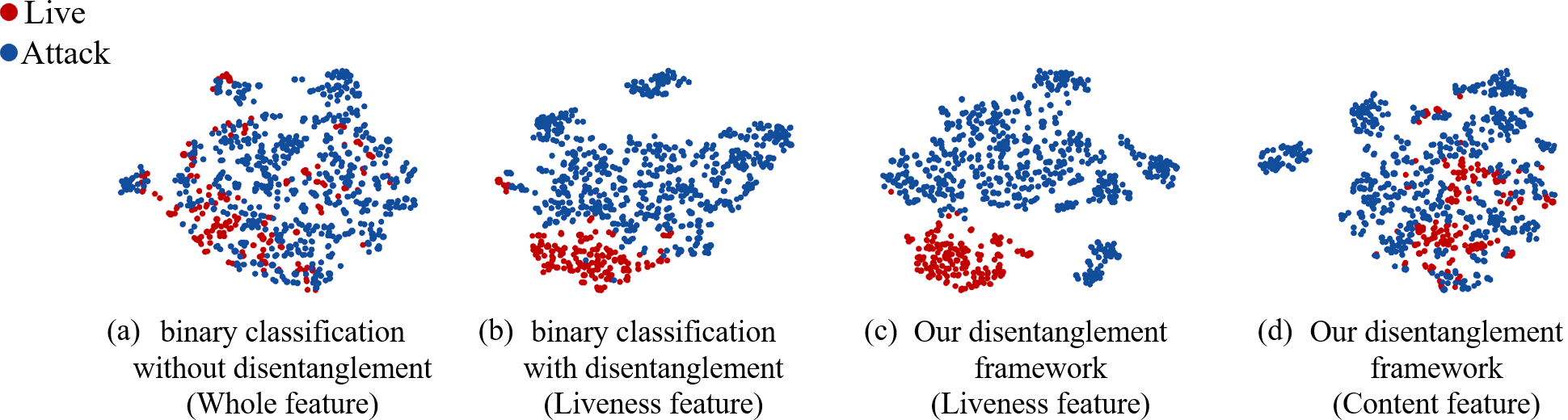}
    \setlength{\belowcaptionskip}{2pt}
    \caption{\textbf{Visualization of feature distributions from different methods}. We use different constraint on livness feature or whole feature and draw the corresponding feature in brackets by t-SNE~\cite{TSNE}.
    }
    \label{ablation_TSNE}
\end{figure}

We use t-SNE~\cite{TSNE} to visualize the features from different methods, which includes $500$ live face images and $2,000$ spoof images, as illustrated in Fig.~\ref{ablation_TSNE}.
Comparing (a) with (b), we conclude that disentanglement indeed finds a sub-space where the features of live and spoof can be distinguished more easily. For comparison between (b) and our method (c), low level LBP supervision on the liveness features improves discrimination between live and attack.
{The difference between (c) and (d) proves that liveness features indeed can distinguish between real and attack while content features can't.}

\begin{table}[t!]
  \centering
  \caption{\textbf{The comparison of different supervision and combination}.}
    \linespread{1.1}\selectfont
         \begin{tabular}{|c|c|c|c|c|c|c|}
         \hline
         \diagbox{ACER}{Method} & BC-Depth & 0/1 Map-Depth & LBP-LBP & Depth-Depth & Depth-LBP & Ours \\ \cline{4-5}
         \hline
         liveness features    & 3.64   & 3.02  & 1.87  & 1.69 & 1.65 & \textbf{1.56} \\
         \hline
         fusion              & 2.78   & 2.50  & 2.40  & 1.80 & 1.50 & \textbf{1.25} \\
         \hline
         \end{tabular}
       \label{LBP_Depth}
\end{table}

\subsubsection{Different Supervision.}a
In our method, we propose the supervision combining low-level LBP texture and high-level depth information. We compare this combination of supervision with other five ablation methods, which are all based on the proposed disentanglement framework:
(1) Binary classification (BC-Depth) method which uses binary classification on the liveness space.
(2) 0/1 Map-Depth method means restricting liveness space by regressing the features to 0/1 Map, where 0 map is for attack and 1 map is for live.
(3) LBP-LBP method supervises feature space and translated images with LBP map. 
(4) Depth-Depth method refers to two depth supervision on feature space and image space.
(5) Depth-LBP method uses depth supervision on feature space and LBP supervision on translated images, which is a reverse version of our method.

Tab.~\ref{LBP_Depth} shows the performance of each method on liveness features and the fusion results with depth network. Compared with different supervison on liveness features, LBP as a low-level texture supervision regularizes the feature space efficiently and performs better. The results of four combinations about LBP and Depth supervision show that the same supervision on feature space and images performs worse than different supervision. And the order of the two supervisions has little effect on the results, but the result of our method is slightly better.

\subsubsection{Score Fusion.}
Using Oulu-NPU protocol 1, we perform studies on the effect of score fusion.
Tab.~\ref{score_fusion} shows the results of each output and the fusion with maximum and average. It shows using LBP map or depth map, the performance is similiar. And the fusion of LBP map and depth map achieves the best performance. Hence, for all experiments, we evaluate the performance by utilizing the fusion score of the LBP map and the depth map, $score = (\left\| map_{lbp} \right\| + \left\| map_{depth} \right\|)/2$.

\begin{figure}[t!]
    \centering
    \includegraphics[width=122mm]{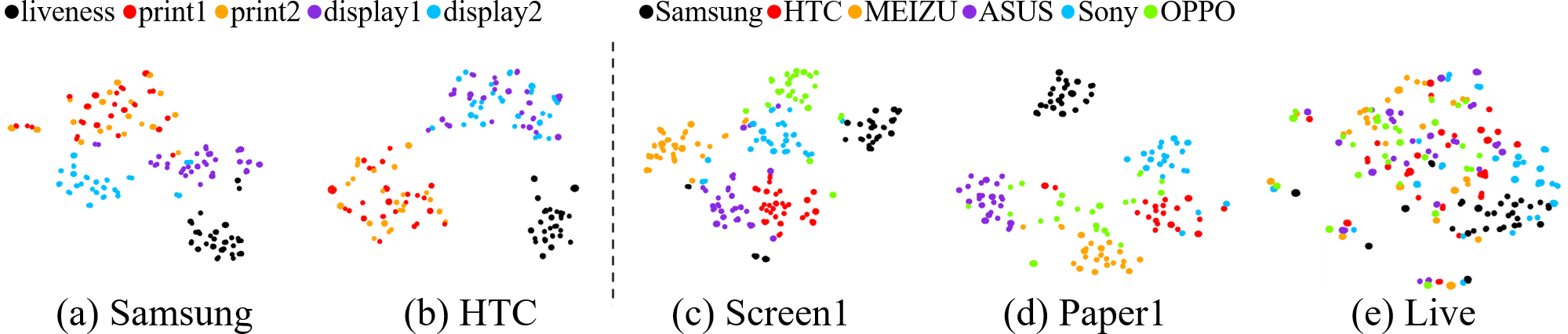}
    \setlength{\belowcaptionskip}{2pt}
    \caption{\textbf{Distribution of liveness features} under two different settings: (a) and (b) display liveness features of different attack and live with the same devices; (c), (d) and (e) are about features of different devices with the same attack or live.}
    \label{spoof_device_type}
\end{figure}

\begin{table}[t!]
  \centering
  \caption{\textbf{The results of score fusion.}}
    \linespread{1.1}\selectfont
      \begin{tabular}{|c|c|c|c|c|}
      \hline
      \multirow{2}*{Method} & \multirow{2}*{LBP Map} & \multirow{2}*{Depth Map} & \multicolumn{2}{c|}{Fusion} \\ \cline{4-5}
             &      &      & Maximum & Average \\
      \hline
      APCER & \text{1.25} & 2.50 & 2.92 & 1.67 \\
      \hline
      BPCER & 1.67 & 0.83 & 0.83 & \text{0.83} \\
      \hline
      ACER  & 1.56 & 1.67 & 1.88 & \text{1.25} \\
      \hline
      \end{tabular}
  \label{score_fusion}
\end{table}

\section{Further Exploration}
We have ruled out the effects of some factors on liveness features in Sec.~\ref{Translation Result}. For better understanding the essence of the liveness features, we do some qualitative experiments to explore what factors are related to it.

\noindent\textbf{Spoof Type.}
We randomly pick up 200 images, which are collected by one certain device. Then we extract the liveness features of images and visualize them by t-SNE~\cite{TSNE}. We demonstrate results under Samsumng and HTC mobiles in Fig.~\ref{spoof_device_type}(a) and (b). Although no additional constraints on attacks are used, there are at least three distinct clusters: live images, paper attack and screen attack in all equipment, which implies liveness features may be related to the spoof type.
\begin{figure}[t!]
    \centering
    \includegraphics[width=122mm]{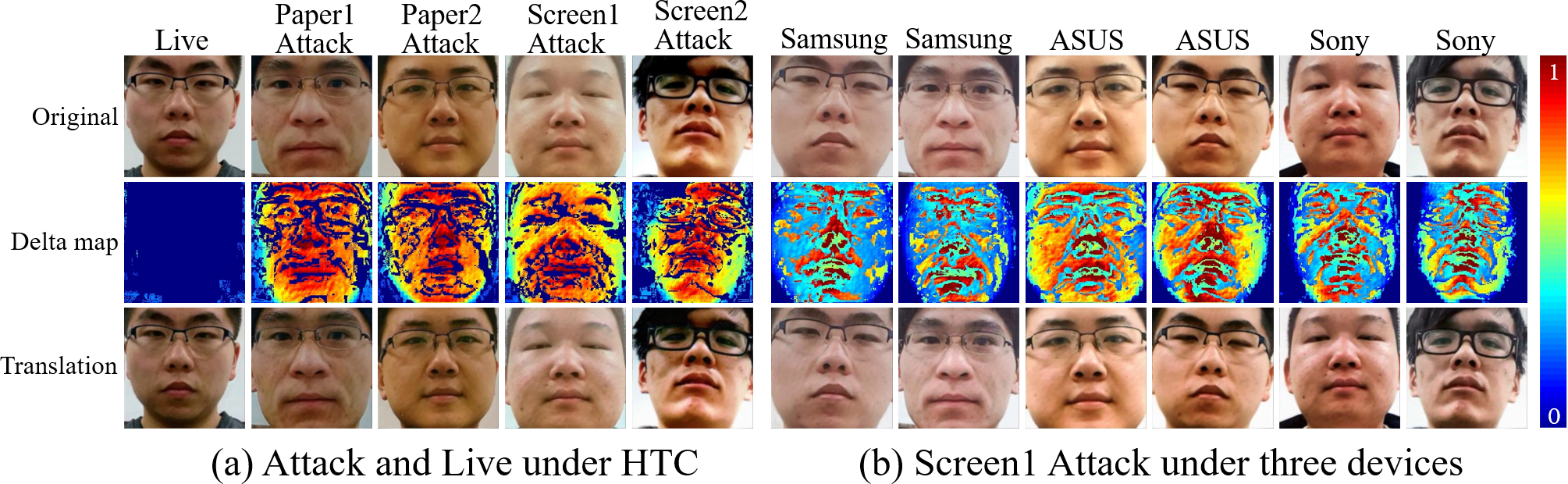}
    \caption{\textbf{The delta maps} for different attacks with same device and different devices with same attack. }
    \label{device_attack_delta}
\end{figure}

\noindent\textbf{Collection Equipment.}
We randomly pick up 200 images for each type of attack and live with six different devices. Then we visualize the liveness features in Fig.~\ref{spoof_device_type}(c), (d) and (e). The liveness features from different devices are clustered for attacks, but scattered for live person. It shows that the liveness features of real person may not related to collection equipment. However, the liveness features of attack may include information on collection equipment.

We further display the pixel-wise delta map between generated images and original images of each type, as shown in Fig.~\ref{device_attack_delta}.
The original images, which are shown in the first row, exchange the liveness features with the same one live image to generate the results in the third row. Then we subtract translation images from original images to get delta maps, which are mapped into color space for a better visualization in the second row.
From Fig.~\ref{device_attack_delta}, we may get the following conclusions:
(1) When exchanging liveness features between real faces, the delta maps are almost zero. However the delta maps become bigger when between live faces and spoof images.
(2) Delta maps of the same type of attack (paper or screen) are similar but are distinguishing between two kinds of attacks.
(3) For the same type of attack, delta maps are different under different collection equipment.

\section{Conclusions}
This paper introduces a new perspective for face anti-spoofing that disentangles the liveness and content features from images. A novel architecture combining the process of disentanglement is proposed with multiple appropriate supervisions. We combine low-level texture and high-level depth characteristics to regularize the liveness space. We visualize the translation process and analyze the content of the liveness features which provides a deeper understanding of face anti-spoofing task.
Our method is evaluated on widely-used face anti-spoofing databases and achieves outstanding results.

\clearpage
\bibliographystyle{splncs04}
\bibliography{3315}
\end{document}